\documentclass[sigconf,nonacm]{aamas}

\usepackage{balance} 
\usepackage{times}
\usepackage{url}
\usepackage[utf8]{inputenc}
\usepackage{graphicx}
\usepackage{amsmath}
\usepackage{booktabs}
\usepackage[linesnumbered,ruled,vlined]{algorithm2e}
\usepackage{nomencl}
\usepackage{wrapfig}
\usepackage{multirow}
\usepackage{tikz}
\usepackage{tabularx}
\usepackage{microtype}

\usepackage{enumitem}
\usetikzlibrary{calc, arrows, positioning, shapes, shadows, spy, plotmarks, matrix, fit, backgrounds}

\acmSubmissionID{<<submission id>>}

\title[AAMAS-2025 Formatting Instructions]{Visual IRL for Human-Like Robotic Manipulation}

\author{Ehsan Asali}
\affiliation{
  \institution{The University of Georgia}
  \city{Athens}
  \country{United States}}
\email{ehsanasali@uga.edu}

\author{Prashant Doshi}
\affiliation{
  \institution{The University of Georgia}
  \city{Athens}
  \country{United States}}
\email{pdoshi@uga.edu}

\begin{abstract}
We present a novel method for collaborative robots (cobots) to learn manipulation tasks and perform them in a human-like manner. Our method falls under the learn-from-observation (LfO) paradigm, where robots learn to perform tasks by observing human actions, which facilitates quicker integration into industrial settings compared to programming from scratch. We introduce \virl{} that uses the RGB-D keypoints in each frame of the observed human task performance directly as state features, which are input to inverse reinforcement learning (IRL). The inversely learned reward function, which maps keypoints to reward values, is transferred from the human to the cobot using a novel neuro-symbolic dynamics model, which maps human kinematics to the cobot arm. This model allows similar end-effector positioning while minimizing joint adjustments, aiming to preserve the natural dynamics of human motion in robotic manipulation. In contrast with previous techniques that focus on end-effector placement only, our method maps multiple joint angles of the human arm to the corresponding cobot joints. Moreover, it uses an inverse kinematics model to then minimally adjust the joint angles, for accurate end-effector positioning. 
We evaluate the performance of this approach on two different realistic manipulation tasks. The first task is produce processing, which involves picking, inspecting, and placing onions based on whether they are blemished. The second task is liquid pouring, where the robot picks up bottles, pours the contents into designated containers, and disposes of the empty bottles. Our results demonstrate advances in human-like robotic manipulation, leading to more human-robot compatibility in manufacturing applications. 
\end{abstract}


\newcommand{\BibTeX}{\rm B\kern-.05em{\sc i\kern-.025em b}\kern-.08em\TeX}
\newcommand{\virl}{{Visual IRL}}
\newcommand{\ns}{{Neuro-Symbolic Dynamics Mapping}}

\begin{document}


\pagestyle{fancy}
\fancyhead{}


\maketitle 


\section{Introduction}
\label{sec:intro}
Several factors are known to contribute to successful and sustainable human-robot collaborations~\cite{lin2021sustainable, ajoudani2018progress}. One of these is whether the collaborative robot is likable. Robots tend to be likable when they exhibit social intelligence and act in ways that make the human partner comfortable. One of these ways is for the robot to use naturalistic, human-like, motions in performing its tasks.   

\begin{figure*}[t]
\setlength{\belowcaptionskip}{-5pt}
\centerline{\includegraphics[width=1.0\textwidth]{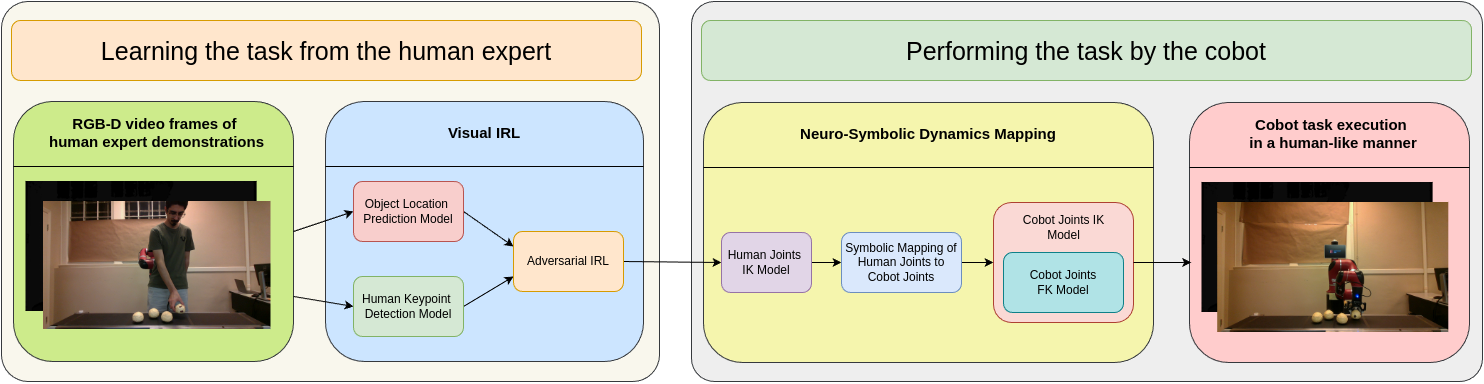}}
\caption{Overview of the proposed method’s pipeline: The pipeline begins by learning the task from a human expert that starts with capturing the RGB-D stream during the human's task performance. Object location prediction and human keypoint detection models extract state parameters for AIRL, which learns a reward function that reflects the expert's preferences. When deployed on the cobot, the learned reward function is used to achieve a policy to perform the task. The calculated policy is input into the Neuro-Symbolic Dynamics Mapping model, which maps human joints to cobot joints and generates initial joint angles for the cobot. These angles are refined using the cobot joints IK model to adjust the end-effector positioning for accurate manipulation, allowing the cobot to perform the task successfully.}
\label{fig:virl_pipeline}
\Description{}
\end{figure*}

Toward achieving this goal, we adopt the learn-from-observation (LfO) paradigm, which is a human-inspired methodology where robots learn to perform tasks by observing humans perform them. This approach can facilitate quicker integration of collaborative robots (cobot) in industrial settings without extensive programming from scratch. In this paper, we present a novel approach for cobots to learn manipulation tasks and execute them in a naturalistic, human-like manner. Our method leverages real-time human pose estimation and object detection on every RGB-D frame of the observed human task performance. Three-dimensional keypoints from these frames are extracted, polished, and then used directly as state features input into an adversarial inverse reinforcement learning (AIRL) technique~\cite{fu2018learning}. This technique learns a reward function that is interpreted as human preferences for the performed task. The learned reward function and corresponding behavior, which maps the pose keypoints to actions in terms of human wrist joint angles, is transferred to a cobot using a novel neuro-symbolic dynamics model. This model maps human kinematics to a cobot arm, allowing for similar end-effector positioning while minimizing joint adjustments, thus preserving the natural dynamics of human motion in robotic tasks. An overview of our proposed method's pipeline is depicted in Fig.~\ref{fig:virl_pipeline}.

Our methodology offers two key contributions. First is the {\em visual IRL} that subscribes to the LfO paradigm and performs IRL directly with human pose keypoints as state features obtained from a state-of-the-art standard real-time pose estimation model such as YOLO~\cite{wang2023yolov7}. We demonstrate that IRL can scale to a stream of such state features and yield an accurate reward function. This mitigates the need to train custom state-action recognition models such as SA-Net~\cite{soans2020sa} or MVSA-Net~\cite{asali2023mvsa} to process video frames and obtain high-level states and actions as input to IRL algorithms. Second is the {\em neuro-symbolic dynamics mapping model}. This model enables the transfer of human arm motion dynamics to a cobot with differing degrees of freedom (DoF), maintaining the human-like characteristics of the motion. Unlike previous techniques that focus solely on end-effector placement, our method maps multiple joint angles of the human arm to corresponding cobot joints. After mapping the joints and calculating the initial joint angles for the cobot, an inverse kinematics model is used to make adjustments for accurate end-effector positioning with minimum changes to the initial joint angles. This approach ensures that the cobot performs the task while adopting the style and fluidity of human movement.

In the Experiments section, we examine this methodology for learning two manipulation tasks with real-world applications and performing each with a different cobot. One task is produce processing which involves picking, inspecting, and placing onions at different locations based on whether the onion is blemished or not~\cite{asali2023mvsa, sengadu2023dec}. Recordings of humans sorting onions in the lab are passed through the method and used to train the 7 DoF Sawyer cobot to perform onion sorting with human-like motion. The second task is liquid pouring which involves picking up colored bottles, pouring contents into a corresponding container (based on the color of the bottles), and disposing of the empty bottles into a bin. A 6 DoF KUKA cobot (LBR iisy) is used to perform the task in a human-like manner. These tasks are chosen due to their relevance to manufacturing automation and inherent complexity due to variations in human judgment and motion.

We rigorously evaluate our method using multiple metrics, including Learned Behavior Accuracy (LBA), Mean Squared Error (MSE), Root Mean Squared Error (RMSE), Mean Absolute Error (MAE), coefficient of determination (R²), and we also measure the averages of end-effector displacement, movement jerkiness, and the time in which it takes to perform the sorting action and compare it with the baseline path planner. The evaluation demonstrates advances in achieving human-like sorting actions, highlighting the robustness and applicability of our method in real-world scenarios.

\noindent We summarize our contributions as follows:

\begin{itemize}[leftmargin=*, itemsep=0.5ex, topsep=0in, parsep=0in, partopsep=0in]
    \item We present an end-to-end perception-based inverse reinforcement learning approach that enables cobots to perform manipulation tasks in a human-like manner.
    \item We develop a novel neuro-symbolic dynamics mapping model that enables transferring human arm motion dynamics to a cobot with different DoFs.
    \item We conduct real-world experiments on two different cobots namely Sawyer and KUKA LBR iisy with 7 and 6 DoFs, respectively, showing improvement in achieving human-like manipulation actions.
\end{itemize}

\section{Related Work}
\label{sec:related}

The \textbf{Learn-from-Observation (LfO)} paradigm enables robots to learn skills by observing human demonstrations, reducing the need for explicit programming. Advances like~\cite{asali2023mvsa} and~\cite{soans2020sa} introduced deep neural networks for state-action recognition from RGB-D streams, enhancing task transfer to robots. Reviews such as~\cite{eze2024learning} and~\cite{ravichandar2020recent} categorize video-based learning methods, discussing strengths and visual data's role in LfO. One-shot imitation learning, explored in~\cite{yu2018one} and~\cite{finn2017one}, leverages domain adaptation to learn tasks from single observations. Studies like~\cite{vakanski2017robot} and~\cite{young2021visual} integrate visual observation with neural networks, simplifying complex task learning from demonstrations.

\textbf{Inverse Reinforcement Learning (IRL)} aims to infer the underlying reward function from observed expert behaviors, facilitating the learning of policies that mimic these behaviors~\cite{Arora19:Survey}. To enhance human-robot teaming, Decentralized Adversarial IRL (Dec-AIRL) utilizes decentralized demonstrations and adversarial training~\cite{sengadu2023dec}. Addressing occlusions and observer noise, MMAP-IRL employs marginal MAP estimation for more reliable reward inference~\cite{Suresh22:Marginal}. Leveraging GANs, the method by Fu et al.~\cite{fu2018learning} enhances the robustness of learned rewards. In model-based IRL, Das et al.~\cite{das2021model} integrate visual perception with IRL to facilitate complex task learning from visual demonstrations. Online IRL approaches, as explored by Arora et al.~\cite{arora2023online}, incorporate real-time observations for continuous adaptation. In another work, Zakka et al.~\cite{zakka2022xirl} introduce Cross-embodiment IRL (XIRL), using temporal cycle consistency to transfer learned rewards across different embodiments.

\textbf{Neuro-Symbolic Models} combine neural networks with symbolic reasoning to leverage the strengths of both methods. A scalable method by Bendinelli et al. employs large-scale pre-training to uncover symbolic equations from data~\cite{biggio2021neural}. Enhancing transparency in learning processes, Gopalan and Reichlin's approach focuses on controllable neural symbolic regression~\cite{bendinelli2023controllable}. The innovative use of symbolic building blocks instead of traditional activations by Martius and Lampert allows for extracting formulas from trained networks~\cite{boddupalli2023symbolic}. Merging neural predictions with symbolic search, Landajuela et al.'s unified framework achieves better generalization~\cite{landajuela2022unified}. Discovering algebraic formulas that explain learned models, Cranmer et al.~\cite{cranmer2020discovering} integrate graph neural networks with symbolic regression. Finally, Udrescu and Tegmark's AI Feynman uses physics-inspired techniques to identify and simplify symbolic expressions, enhancing model interpretability~\cite{udrescu2020ai}.

\textbf{Mapping human dynamics to robots} involves capturing, analyzing, and replicating human motion patterns to enable robots to perform tasks with human-like precision. Dube and Tapson~\cite{dube2009kinematics} developed a 10-DoF arm design to improve the kinematic range using visual motion capture. Memmesheimer et al.~\cite{memmesheimer2019simitate} introduced Simitate, focusing on end-effector motion using human keypoints, differing from our full-arm approach. AVID translates human videos into robotic actions with deep learning for multi-stage tasks~\cite{smith2019avid}. Sena and Howard~\cite{sena2020quantifying} emphasized capturing detailed human motion data, e.g. joint angles and velocities to enhance robot learning from demonstrations. Shao et al.~\cite{shao2021concept2robot} presented Concept2Robot, which integrates demonstrations and language instructions for teaching complex tasks. Yamada et al.~\cite{yamada2023efficient} focused on skill acquisition in obstacle-rich environments. Mandikal et al.~\cite{mandikal2022dexvip} proposed DexVIP, enhancing robotic grasping using human hand pose priors from videos.

\textbf{Retargeting Approaches.}
Human-to-robot motion retargeting has been extensively studied in the robotics community. 
For example, Wang et al.~\cite{wang2017generative} proposed a generative retargeting framework using only a single depth sensor, focusing on adapting human poses to robots with similar skeletal structures.
Darvish et al.~\cite{darvish2019whole} explored a geometric retargeting method specifically for whole-body humanoid robots, often relying on motion capture systems.
More recently, Qin et al.~\cite{qin2023anyteleop} presented AnyTeleop for dexterous arm-hand teleoperation with a vision-based approach. 
These methods often emphasize direct pose replication, sometimes requiring specialized hardware or focusing on partial kinematic structures such as hands. 
In contrast, our neuro-symbolic method targets general cobot arms with varying degrees of freedom (DoF), preserving human-like arm dynamics for the entire manipulation task without specialized sensors beyond an RGB-D camera.


\section{Background}
\label{sec:background}

Inverse reinforcement learning (IRL) is concerned with determining the reward function in a Markov decision process (MDP) \cite{puterman2014markov} that best explains the observed behavior of an expert. An MDP is represented by the tuple $\langle S, A, T, R, \gamma, \rho_0 \rangle$, where $S$ is the set of states, $A$ is the set of actions, $T: S \times A \times S \to [0,1]$ defines the state transition probabilities, $R: S \times A \to \mathbb{R}$ is the reward function, $\gamma \in [0,1)$ is the discount factor, and $\rho_0$ is the initial state distribution. In the IRL context, the learner knows the states $S$, actions $A$, discount factor $\gamma$, initial state distribution $\rho_0$, and state transition probabilities $T$, but the reward function $R$ is unknown. Model-free approaches can forgo the need for $T$. The policy $\pi: S \to A$ is a function that maps states to actions. Let $X^E$ represent the set of expert demonstrations; a complete trajectory $X$ can be denoted as $X = (s_1, a_1, s_2, a_2, \ldots, s_T, a_T)$, with $X \in X^E$. 

The objective of IRL is to estimate the reward function $R$, which, when combined with the known components of the MDP $\langle S, A, T, \gamma, \rho_0 \rangle$, induces a policy $\pi$ that produces trajectories similar to the expert demonstrations $X^E$. These expert trajectories consist of a sequence of state-action pairs $(s_t, a_t)$, where $s_t \in S$ and $a_t \in A$ for each time step $t$. By observing the expert's transitions from state $s_t$ to state $s_{t+1}$ under action $a_t$, the learner infers which rewards $R(s_t, a_t)$ make the expert's behavior optimal. The inferred policy $\pi$ is thus expected to maximize the cumulative discounted reward $\sum_{t=1}^T \gamma^t R(s_t, a_t)$ over the trajectory, where $\gamma$ determines how future rewards are weighed relative to immediate rewards.

Adversarial Inverse Reinforcement Learning (AIRL) is an approach that builds upon the framework of Generative Adversarial Networks (GANs) to solve the IRL problem by simultaneously learning a policy and a reward function. AIRL is based on the maximum causal entropy IRL framework \cite{ziebart2010modeling}, which considers an entropy-regularized MDP. The goal is to find the optimal policy $\pi^*$ that maximizes the expected entropy-regularized discounted reward:

\begin{equation}
\pi^* = \arg\max_{\pi} {E}_{\tau \sim \pi} \left[ \sum_{t=0}^{T} \gamma^t (R_\theta(s_t, a_t) + \mathcal{H}(\pi(\cdot|s_t))) \right]
\end{equation}

where $\tau = (s_0, a_0, \ldots, s_T, a_T)$ denotes a trajectory, and $\mathcal{H}(\pi(\cdot|s_t))$ represents the policy's entropy at state $s_t$. AIRL frames the IRL problem as a GAN optimization \cite{goodfellow2014generative}, with the discriminator $D_\theta(s, a)$ distinguishing between expert demonstrations and generated trajectories:

\begin{equation}
D_\theta(s, a) = \frac{\exp(f_\theta(s, a))}{\exp(f_\theta(s, a)) + \pi(a|s)}
\end{equation}

where $f_\theta(s, a)$ approximates the advantage function. At optimality, $f_\theta(s, a) = \log \pi^*(a|s) = A^*(s, a)$. The reward function is updated based on the discriminator's output:

\begin{equation}
R_\theta(s, a) \leftarrow \log D_\theta(s, a) - \log (1 - D_\theta(s, a)).
\end{equation}

The policy $\pi$ is optimized with respect to the learned reward function $R_\theta$ using any policy optimization method, such as Trust Region Policy Optimization (TRPO) \cite{schulman2015trust}. The iterative process of AIRL ensures the reward function is disentangled from the environment dynamics, leading to better generalization across different tasks and environments \cite{fu2018learning}.

\section{Method}
\label{sec:method}

This work presents a novel approach that addresses the challenge of teaching cobots to perform manipulation tasks in a human-like manner. We first introduce \virl{} which intends to learn the task preferences using human body keypoints and object location data. Then, we present the \ns{} model that intelligently translates the human arm's motion dynamics to a cobot with a different DoF. This model enables the cobot to apply the learned policy with joint motions similar to a human subject.

\subsection{\virl{}}
In order to learn the human's intentions as well as human kinematics, we suggest a novel IRL method, called \virl{}. \virl{} uses human arm keypoints as part of its state and action parameters to learn the human kinematics and takes advantage of AIRL to learn the task preferences. Moreover, it utilizes the 3D location of the object of interest to discover human intentions further. \\

\noindent\textbf{Problem Formulation.} In the Visual Inverse Reinforcement Learning (Visual IRL) method, we model the problem as a single-agent Markov Decision Process defined as \(\text{MDP} = \langle S, A, T, R \rangle\) situated in a Cartesian space. The state space (\(S\)) includes the end-effector location (\(eef_{loc}\)), object location (\(onn_{loc}\)), and object label prediction (\(onn_{pred}\)), all in 3D space. The action space (\(A\)) comprises changes in the end-effector's 3D coordinates, represented as \(\Delta(x, y, z)\). The reward function (\(R\)) assigns positive values for optimal manipulation actions. The transition function (\(T\)) models the probability of transitioning between states given specific actions, defined as \(T: S \times S \times A \rightarrow [0, 1]\). The goal is to learn both a policy (\(\pi\)) that maps states containing human keypoints to actions involving the end-effector's movements and a reward function (\(R\)) that reflects the human's preferences, denoted as \(\pi_{H}: S_{H} \rightarrow A_{H}\) and \(R: S \times A \rightarrow \mathbb{R}\), respectively.\\

\begin{figure}[!ht]
\vspace{-9mm}
\centering
\begin{tikzpicture}[thick, spy using outlines={rectangle,lens={scale=2.5}, width=1cm, height=1.5cm, connect spies}]
	\node (reg_id1) {\includegraphics[width=0.9\columnwidth]{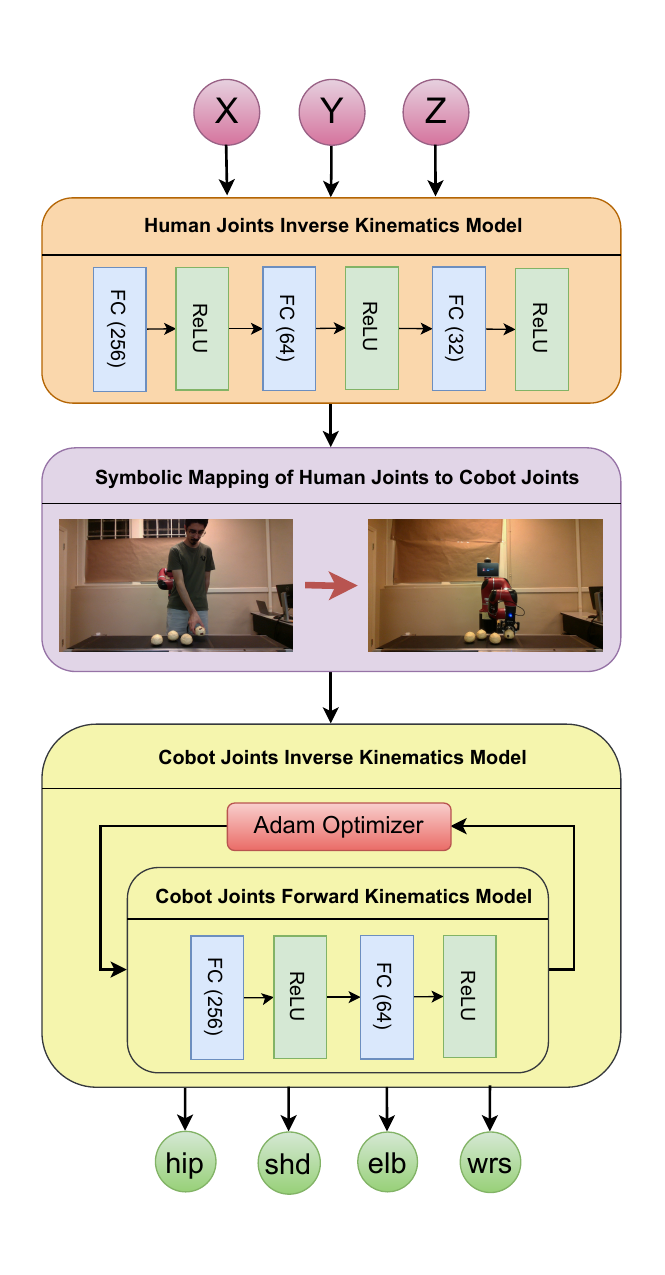}};
\end{tikzpicture}
\vspace{-12mm}
\caption{An overview of our Neuro-Symbolic Dynamics Mapping architecture. The wrist's 3D coordinates are input into the human joint IK model to obtain human joint coordinates. These coordinates are then converted into cobot joint angles through symbolic mapping. The initial cobot joint angles are iteratively refined using the cobot's FK and IK models. Once the joint angles meet the desired threshold, they are finalized and used by the robot to reach the target in a human-like manner.}
\label{fig:ns_arch}
\Description{}
\end{figure}

\noindent\textbf{AIRL Implementation.}
Although Section~\ref{sec:background} outlined AIRL conceptually, here we highlight how \virl{} obtains the reward function $R(s,a)$ and policy $\pi$ from human demonstrations. 
Specifically, we collect expert trajectories $(s_t, a_t)$ derived from pose keypoints and object localization, and feed these into the AIRL framework. 
Within AIRL, a discriminator $D_\theta(s,a)$ (Eq. 2) distinguishes expert from generated trajectories, thereby refining the learned reward:
\begin{equation}
R_\theta(s, a) \leftarrow \log D_\theta(s, a) - \log (1 - D_\theta(s, a)).
\end{equation}
Subsequently, we apply TRPO to optimize a policy $\pi(a|s)$ that maximizes the cumulative reward. 
Thus, IRL in our pipeline is not merely replicating poses but inferring task-specific preferences from demonstrations, ensuring that the cobot’s learned behaviors align with expert intentions \emph{and} human-like motion dynamics.

\noindent\textbf{Human Keypoint Detection.} In order to learn the human arm motion dynamics, the human body skeleton should be detected and tracked at every time step. In this regard, YOLOv8 pose estimation model is used as the base model for 2D human keypoint detection. The 2D output coordinates get converted to 3D using the camera's internal and external intrinsic information. While the two models are highly accurate in most cases, situations may arise where one or both of them fail to detect joint locations due to occlusions, rapid movements, or sensor noise. To address this issue, we suggest a keypoint prediction model that can estimate the 3D coordinates of the human arm and hip joints \((hip_{H'}^{t}, shd_{H'}^{t}, elb_{H'}^{t}, wrs_{H'}^{t})\), when pose estimation predictions are unreliable. The input features include the current (\(t\)) and previous (\(t-1\)) 3D coordinates of the hip, shoulder, elbow, wrist, index finger, and thumb joints, represented as \((hip_{H}^{t}, shd_{H}^{t}, elb_{H}^{t}, wrs_{H}^{t})\) and \((hip_{H}^{t-1}, shd_{H}^{t-1}, elb_{H}^{t-1},\\wrs_{H}^{t-1})\), respectively.

The network architecture comprises two LSTM layers each containing 64 units, followed by two fully connected layers with 256 and 64 units using ReLU activations. The output layer predicts the 3D coordinates for the joints, resulting in an output size of 12 including the predicted 3D coordinates of the six input joints. The model is trained using the MSE loss function and the Adam optimizer with a learning rate of 0.001.\\

\noindent\textbf{Object Localization and Prediction}. This model aims to find the 3D location of the object to be sorted. Initially, the custom YOLOv8 object detection model provides 2D bounding box coordinates and labels for all of the objects of interest in the current frame. This 2D data is converted to 3D using the camera's intrinsic parameters. Similar to pose estimation models, the YOLOv8 object detection model might lose track of objects due to occlusions or rapid movements of the human or robot. For such scenarios, we propose a model that uses the previous 3D location of the object (\(obj_{loc}^{t-1}\)) and the previous (\(eef_{loc}^{t-1}\)) and current (\(eef_{loc}^{t}\)) locations of the end-effector to predict the current 3D location of the object (\(obj_{loc}^{t}\)):
\begin{equation}
{obj}_{loc}^{t} = f(obj_{loc}^{t-1}, eef_{loc}^{t-1}, eef_{loc}^{t}).
\end{equation}

The model's architecture includes an input layer processing sequences of 3D coordinates, followed by two LSTM layers with 64 and 32 units, and two fully connected layers with 256 and 64 units using ReLU activations. The output layer provides adjusted 3D coordinates of the object. The model ensures accurate localization by leveraging spatial and temporal data from previous predictions and current detections, enhancing the reliability of the pipeline.

\subsection{Neuro-Symbolic Human-to-Robot Dynamics Mapping}
As part of our framework, we propose a \ns{} method to effectively transfer human motion dynamics to a robot. This method ensures accurate end-effector placement while minimizing joint adjustments, thereby preserving the natural dynamics of human motion in robotic tasks such as pick-and-place operations. As an example, the process of mapping the human kinematics to Sawyer (which is a 7 DoF cobot) is shown in Fig.~\ref{fig:ns_arch}. This method involves multiple chronological steps, explained below in detail in ascending order.\\

\noindent\textbf{Human Joints Inverse Kinematics Model.} 
To transfer the human kinematics to the cobot, the wrist joint can be used as the reference point and other joints can be aligned accordingly. The human joints inverse kinematics model predicts the 3D coordinates of the human hip, shoulder, and elbow joints based on the 3D location of the wrist. The model takes the 3D coordinates of the wrist at each time step and predicts the 3D coordinates of the other joints for the same time step. Although the training data may contain multiple samples with similar wrist locations yet slightly different human poses, the inverse kinematics model averages these variations to determine a representative human pose. The model consists of three fully connected layers with 256, 64, and 32 units using ReLU activations, followed by an output layer predicting the 3D coordinates of the three joints, resulting in a total of 9 output labels. The model is trained using the Mean Squared Error (MSE) loss function and the Adam optimizer with a learning rate of 0.001. \\

\begin{figure*}[t]
\setlength{\belowcaptionskip}{-5pt}
\centerline{\includegraphics[width=0.8\textwidth]{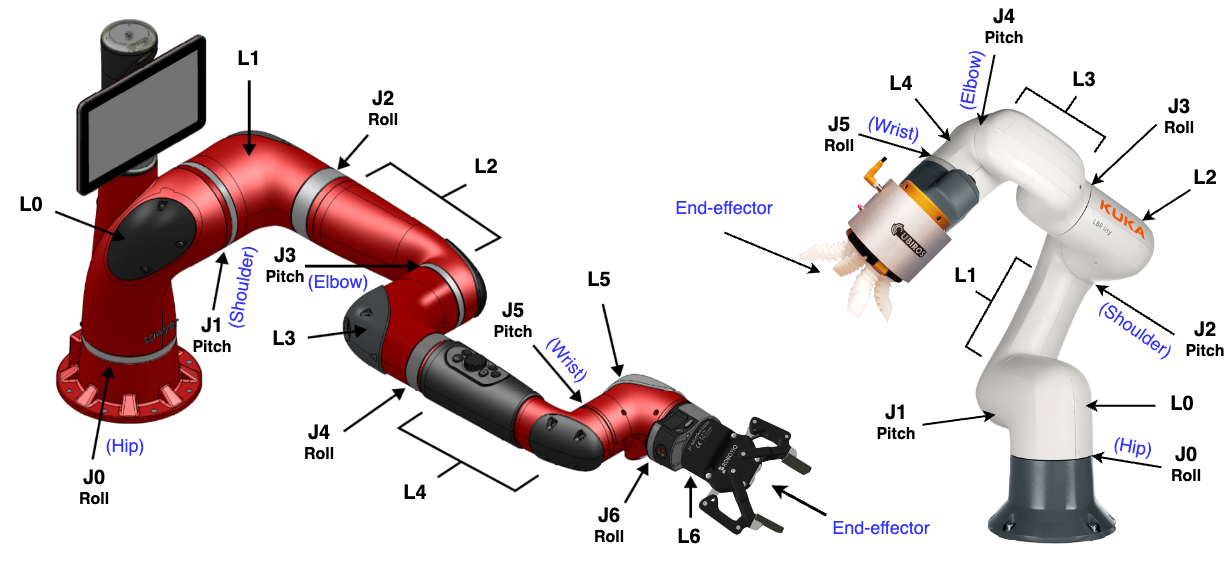}}
\caption{One-to-one mapping of Sawyer and LBR iisy KUKA joints to the human joints. The human joints are denoted in blue color.}
\label{fig:sawyer_kuka}
\Description{}
\end{figure*}

\noindent\textbf{Symbolic Mapping of Human Joints to Cobot Joints.} 
This symbolic model maps human joint angles (hip, shoulder, elbow, and wrist) to the cobot's joint angles. The one-to-one mapping of human joints and the joints of the two cobots used in our experiments is depicted in Fig.~\ref{fig:sawyer_kuka}. The input to the model consists of human joint angles, and the output is the initial robot joint angles. This mapping provides an initial estimate of the robot's joint configuration based on human motion, which is the starting point for the subsequent neural models.\\


\noindent\textbf{Cobot's Restricted Forward Kinematics Model.} 
The cobot's restricted forward kinematics (FK) model aims to predict the 3D position of the cobot's end-effector based on the angles of the hip, shoulder, elbow, and wrist joints. This model is restricted by assuming that only the hip (\(hip\)), shoulder (\(shd\)), elbow (\(elb\)), and wrist (\(wrs\)) joints of the cobot are moving, while other joints remain stationary (fixed at 0 degrees). The motivation here is to maximize the similarity of the cobot's arm posture to the human since the robot has extra joints. The input to the model consists of the cobot's hip, shoulder, elbow, and wrist angles, and the output is the 3D position of the end-effector (x, y, z). The model architecture includes an input layer accepting the joint angles, followed by two fully connected layers with 256 and 64 units using ReLU activations. The final output layer predicts the 3D coordinates of the end-effector. The model is trained using the Mean Squared Error (MSE) loss function and the Adam optimizer with a learning rate of 0.001. The training data is collected by running the cobot on Gazebo and Rviz simulations, recording the joint movements and corresponding end-effector positions.\\

\noindent\textbf{Cobot's Inverse Kinematics Model.}
This model is designed to optimize the joint angles of the robot's hip (\(hip\)), shoulder (\(shd\)), elbow (\(elb\)), and wrist (\(wrs\)) to ensure that the robot's end-effector reaches the desired target location while maintaining a human-like pose. The model starts with an initial guess for the joint angles and iteratively uses the trained forward kinematics model to estimate the end-effector's position based on the updated angles. The inputs to the model are the initial robot's hip, shoulder, elbow, and wrist angles (\(hip\), \(shd\), \(elb\), \(wrs\)), and the target end-effector position in 3D space (x, y, z). The output of the model is the adjusted robot's hip, shoulder, elbow, and wrist angles (\(hip'\), \(shd'\), \(elb'\), \(wrs'\)). 

The optimization process involves minimizing the distance between the estimated end-effector location and the desired location while also minimizing changes to the initial joint angles. The model is trained using the Mean Squared Error (MSE) loss function, and the optimization is performed using the Adam optimizer with a learning rate of 0.01. The optimization objective includes two components in the loss function: position error and adjustment penalty. The position error measures the difference between the predicted end-effector position and the desired position, while the adjustment penalty measures the deviation of the adjusted joint angles from the initial joint angles. 

The loss function \( L \) can be formulated as follows:
\begin{equation}
L = \| \hat{p}_{ee} - p_{ee} \| + \alpha \| \theta - \theta_0 \|
\end{equation}
where \( \hat{p}_{ee} \) is the estimated end-effector position, \( p_{ee} \) is the desired end-effector position, \( \theta \) represents the current joint angles, \( \theta_0 \) represents the initial joint angles, and \( \alpha = 0.0005 \) is a weight parameter balancing the importance of position accuracy and minimal adjustment.

The optimization proceeds iteratively for a maximum of 10,000 iterations, with the learning rate decaying by a factor of 0.9 every 1,000 steps. The process converges when the position error is below a specified threshold (0.01 m). The model also employs gradient clipping with a maximum norm of 1.0 to stabilize the optimization process. The optimized joint angles (\(hip'\), \(shd'\), \(elb'\), \(wrs'\)) are obtained when the model converges, ensuring the cobot's end-effector reaches the desired target location while maintaining a human-like pose.

\noindent
\textbf{Rationale for Restricted Kinematics.}
Standard URDF-based forward and inverse kinematics typically find \emph{any} valid configuration matching the end-effector pose, potentially activating joints that do not mirror human arm motions. 
In contrast, our Restricted FK model (Section~\ref{sec:method}) fixes non-human-like joints, allowing only those (e.g.\ hip, shoulder, elbow, wrist) that correspond to human arm segments to move. 
Likewise, our specialized IK model jointly minimizes pose error \emph{and} deviations from the initial human-like angles. 
This ensures that the resulting solution not only reaches the target position but also preserves the human motion dynamics captured from demonstrations. 
Such trainable, domain-specific kinematics models are flexible across different robot DoFs, which standard URDF solvers alone do not directly offer, especially when the human arm’s joint mapping does not correspond one-to-one with a robot’s entire kinematic chain.\\

\section{Experiments}
\label{sec:experiments}

We demonstrate the performance of the proposed model by evaluating it in two applicable real-world robotic manipulation domains. The targeted tasks involve the robotic automation of processing line tasks. We start by providing a brief description of the tasks followed by our experimentation procedures and results.

\subsection{Domain Specifications}
\label{sec:domain}
\noindent\textbf{Domain Overview.} We first evaluate our proposed model on a \textbf{line sorting} task where the physical cobot Sawyer (from Rethink Robotics) is tasked with observing and learning how a human sorts an arbitrary number of onions (i.e., separating blemished onions from unblemished ones). The human expert looks at the onions on the conveyance and constantly tries to detect whether onions are blemished or not while sorting them. If an onion is detected as blemished while being on the conveyance, it will be picked and discarded into a bin positioned next to the conveyor belt. Otherwise, the human picks up the onion and meticulously inspects it. If the onion is recognized as blemished after inspection, it will be discarded into the bin. Conversely, if it is deemed unblemished, the expert returns it to the corner of the conveyance and proceeds to inspect the next onion until all onions are processed. 

The second task used to evaluate our model is \textbf{liquid pouring} where the physical cobot LBR iisy (from KUKA Robotics) performs the observation and learning of how a human expert pours the contents of different colored bottles into designated containers. In this setup, we have an arbitrary number of blue and red bottles while there are only two containers with blue and black colors. The contents of blue bottles should be transferred into the blue container while the liquid in red bottles should be poured into the black container. After depleting each bottle, regardless of its color, the empty bottle should be dropped into a bin. In this setup, the focus is on robust pick-and-place style manipulations: grasping bottles, aligning them with the designated containers, and pouring. The pipeline does not explicitly model liquid flow or track pouring volume in real-time. Instead, once the cobot’s end-effector aligns the bottle with the container, the actual pouring action is triggered by a simple tilt. \\

\noindent\textbf{Domain Configuration.}
In our setting, both cobots utilize an Intel RealSense L515 RGB-D camera positioned in front of the human expert and the conveyor belt. The conversion from a 2D pixel location $(x, y)$ to 3D world coordinates involves two steps. First, the pixel coordinates are converted to camera coordinates using depth data and camera intrinsics:

\[
X = \frac{(x - c_x) \cdot z}{f_x}, \quad Y = \frac{(y - c_y) \cdot z}{f_y}, \quad Z = \frac{z}{1000}.
\]
Then, the camera 3D coordinates are transformed into the cobot's 3D world coordinates $(Y_{cobot}, Z_{cobot}, X_{cobot})$, based on the relative spatial location of the camera with respect to the cobot's world origin.
The camera position remains stationary throughout the experiment to simplify the 2D to 3D coordinate conversion process.\\

\noindent\textbf{State and Action Parameters.} In order to learn the human expert's task preferences using \virl{} and transfer this knowledge to the cobot to perform the task in a human-like manner, we have formulated both tasks as an MDP problem. 

The \textbf{state} can be adequately represented by three variables: the predicted label and the 3D spatial location of the object of interest and the 3D spatial location of the end-effector. The 3D spatial locations are continuous values and the status of each object can be labeled with respect to the task. For instance, in the line sorting, the status of the onion is either $blemished$, $unblemished$, or $unknown$; while in the liquid pouring task, the status of the bottle is either $red$, $blue$, or $unknown$. The \textbf{action} is defined as the 3D spatial displacement of the end-effector in two consecutive time steps. 

Multiple objects (e.g. onions or bottles) usually appear in a single frame. To focus our object location prediction model on the object currently being manipulated, we establish a set of conditions to achieve the desired outcome. 
Before assessing these conditions, the object detection results for the current frame are arranged in ascending order based on their three-dimensional Y coordinates. Subsequently, each candidate bounding box and its predicted label are taken into consideration: 
\begin{itemize}[leftmargin=*, topsep=0in, itemsep=0in]
  \item If no object is detected with classification confidence of 0.5 or higher, the label of the object is marked as $unknown$, and minus infinity is assigned to its 3D coordinates.
  \item If the Euclidean distance between the object and the end-effector is less than 20 centimeters, the object's label and 3D coordinates of its bounding box centroid are returned. 
  \item Otherwise, the 3D coordinates and the label of the object with the lowest Y value are chosen, disregarding the rest. 
\end{itemize}


\subsection{Baseline}
\label{sec:baseline}
We assess our proposed method's ability to transfer the behavior to the cobot in comparison to two well-known path planners, namely RRT and RRT-connect. The \textbf{Rapidly-exploring Random Tree (RRT)} algorithm~\cite{lavalle2001randomized} operates directly in the cobot's workspace. The goal of RRT is to find a collision-free path by sampling random points in the Cartesian space and incrementally building a tree rooted at the start position. Each new point is connected to the closest node in the tree, extending towards the target while avoiding obstacles. At every time step, the target end-effector location is calculated using the current end-effector location and the learned policy's action (end-effector displacement). The target location is then passed to RRT to find the desired joint angles to reach a certain threshold near the target. The \textbf{RRT-connect} algorithm~\cite{kuffner2000rrt} builds upon RRT by growing two trees simultaneously—one from the start and one from the goal—aiming to connect them. This bidirectional approach allows for faster exploration and pathfinding as the trees extend toward each other, leading to quicker convergence compared to standard RRT.


RRT is used as the baseline for the onion-sorting task and RRT-connect is the baseline for liquid pouring and their performance is compared with our proposed neuro-symbolic dynamics mapping model. We also compare two state-of-the-art object detection (YOLOv8 and Faster-RCNN) and two pose estimation (Mediapipe and YOLOv8) models to find the ones with the highest performance to be used as part of our methodology.

\subsection{Evaluation Criteria}
\label{sec:eval_criteria}

This subsection outlines the evaluation criteria used to assess the performance of the proposed method. We employ several standard statistical metrics including Mean Squared Error (MSE), Root Mean Squared Error (RMSE), Mean Absolute Error (MAE), and R-squared (R\(^2\)) which are widely recognized in statistical analysis for evaluating the performance of machine learning models \cite{chicco2021coefficient}.

In addition to these metrics, we use the following criteria to compare our method with the baselines:

\begin{itemize}
    \item \textbf{Learned Behavior Accuracy (LBA)}: Commonly used in IRL, LBA is computed as the number of demonstrated state-action pairs that match between using the true and learned policies expressed as a percentage of the former which evaluates how accurately the model replicates observed behaviors.\cite{Arora19:Survey}

    \item \textbf{Average Sorting Time}: The mean time required to perform manipulation for a single object, serving as an efficiency metric.

    \item \textbf{Average End-Effector Distance}: The mean distance traveled by the robotic arm's end-effector during manipulation of a single object, reflecting operational efficiency.

    \item \textbf{Average Movement Jerkiness}: The mean angular movement of robotic joints during manipulation of a single object, with lower values indicating smoother, more human-like movements. Jerkiness is sampled every 100 milliseconds and summed for each sorting task.
\end{itemize}

\subsection{Formative Evaluation}

For each of the two tasks, we performed 10 trials with our proposed model and 10 other trials with the baseline model replaced with our neuro-symbolic model. Each trial involves 4 objects to be manipulated, resulting in 40 individual onion manipulations for each of the two models. We gather the data from all these trials to analyze the performance of different parts of our proposed model and we compare the efficiency of our model with the baselines.

\begin{figure*}[t]
\setlength{\belowcaptionskip}{-5pt}
\centerline{\includegraphics[width=1.0\textwidth]{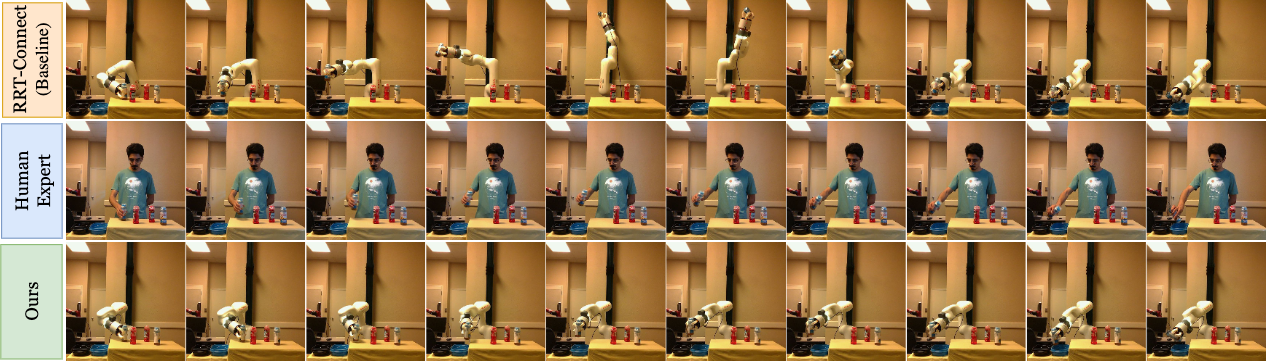}}
\caption{Comparison of motion dynamics between RRT-connect, our proposed model, and human expert. The proposed model closely aligns with human motions, while the baseline exhibits irregularities.}
\label{fig:traj}
\Description{}
\end{figure*}

As mentioned in section IV.A., the human keypoint detection model consists of two major steps: pose estimation and keypoint prediction. In order to select the most accurate pose estimation model, we focused on two state-of-the-art models — Mediapipe and YOLOv8 — and evaluated their performance on 3000 frames of a human expert performing the onion-sorting task. In our Neuro-Symbolic Dynamics Mapping model, four keypoints — the right hip, right shoulder, right elbow, and right wrist — are particularly important as they are mapped to the robot's arm joints for transferring the human manipulation style to the cobot. As shown in Table~\ref{tab:mediapipe_yolo_comparison}, YOLOv8 outperformed Mediapipe in terms of accuracy and consistency, especially in predicting the elbow and wrist keypoints, which are vital for tasks requiring precise movements. YOLOv8’s lower variance provides more reliable inputs for our inverse kinematics model, enabling smoother and more accurate robotic movements. Conversely, Mediapipe's higher variability introduces potential inaccuracies that can hinder fine manipulation tasks. For these reasons, we selected YOLOv8 as the pose estimation model for our keypoint detection pipeline.

\begin{table}[ht!]
\fontsize{7.0}{6.5}\selectfont
\addtolength{\tabcolsep}{-4pt}
\setlength\extrarowheight{3pt}
\caption{Comparison of Mediapipe and YOLOv8's pose estimation accuracy (with the standard deviation) for selected joints.}
\begin{center}
\begin{tabular}{c|cccc}
\hline
{\textbf{Method}} & \textbf{Right Hip} & \textbf{Right Shoulder} & \textbf{Right Elbow} & \textbf{Right Wrist} \\
\hline
\textbf{Mediapipe} & 97.75 $\pm$ 14.832 & 97.92 $\pm$ 13.186 & 84.472 $\pm$ 36.222 & 63.75 $\pm$ 48.079 \\
\textbf{YOLOv8} & \textbf{98.88 $\pm$ 2.492} & \textbf{99.45 $\pm$ 0.635} & \textbf{95.08 $\pm$ 4.643} & \textbf{96.56 $\pm$ 2.886} \\
\hline
\end{tabular}
\label{tab:mediapipe_yolo_comparison}
\end{center}
\end{table}

The human keypoint detection model, assisting in modifying human pose estimation data, is trained on 3000 RGB-D frames and evaluated on 500 RGB-D frames, both randomly extracted from the trial data. The evaluation results are shown in Table~\ref{tab:human_keypoint}.

\begin{table}[ht!]
\centering
\caption{Evaluation results for human keypoint detection model.}
\setlength{\extrarowheight}{2pt} 
\scalebox{0.85}{ 
\begin{tabular}{c|c|cccc}
\hline
{\multirow{2}{*}{\textbf{Joint}}} & {\multirow{2}{*}{\textbf{Axis}}} & \multicolumn{4}{c}{\textbf{Metrics}}\\
& & \textbf{MSE} & \textbf{RMSE} & \textbf{MAE} & \textbf{R\(^2\)} \\
\hline
 & X & 0.000088 & 0.009383 & 0.007238 & 0.579554 \\
Hip & Y & 0.000091 & 0.009546 & 0.008193 & 0.897867 \\
 & Z & 0.000093 & 0.009647 & 0.007862 & 0.289432 \\
\hline
 & X & 0.000077 & 0.008778 & 0.006839 & 0.722046 \\
Shoulder & Y & 0.000062 & 0.007902 & 0.005956 & 0.975333 \\
 & Z & 0.000068 & 0.008217 & 0.006600 & 0.591279 \\
\hline
 & X & 0.000077 & 0.008778 & 0.006839 & 0.722046 \\
Elbow & Y & 0.000053 & 0.007280 & 0.005267 & 0.890235 \\
 & Z & 0.000071 & 0.008426 & 0.006724 & 0.611487 \\
\hline
 & X & 0.000072 & 0.008485 & 0.006001 & 0.987320 \\
Wrist & Y & 0.000053 & 0.007280 & 0.005267 & 0.890235 \\
 & Z & 0.000068 & 0.008217 & 0.006600 & 0.591279 \\
\hline
\end{tabular}
}
\label{tab:human_keypoint}
\end{table}

The object localization and prediction model has to be performed precisely using an accurate object detection model and an auxiliary model to further improve the predicted object location accuracy. We trained Faster-RCNN and YOLOv8 models on 3000 frames and evaluated them using 500 frames. We use Precision, Recall, mAP50, and mAP50-95 metrics to evaluate the two methods. As shown in Table~\ref{tab:faster_yolo_comparison}, YOLOv8 outperforms Faster-RCNN in all criteria.

\begin{table}[ht!]
\fontsize{7.0}{6.5}\selectfont
\addtolength{\tabcolsep}{-4pt}
\setlength\extrarowheight{3pt}
\caption{Comparison of Faster-RCNN and YOLOv8 object detection models in four different criteria.}
\vspace{-2mm}
\begin{center}
\begin{tabular}{c|cccc}
\hline
{\textbf{Method}} & \textbf{Right Hip} & \textbf{Right Shoulder} & \textbf{Right Elbow} & \textbf{Right Wrist} \\
\hline
\textbf{Faster-RCNN} & 92.88 $\pm$ 6.993 & 89.02 $\pm$ 7.577 & 90.75 $\pm$ 6.818 & 69.226 $\pm$ 17.424 \\

\textbf{YOLOv8} & \textbf{98.31 $\pm$ 3.063} & \textbf{91.38 $\pm$ 5.002} & \textbf{95.77 $\pm$ 4.904} & \textbf{82.273 $\pm$ 8.775} \\
\hline
\end{tabular}
\label{tab:faster_yolo_comparison}
\end{center}
\end{table}

The object location prediction model is trained and evaluated on a similar dataset as the object detection models, with the results presented in Table~\ref{tab:object_location}. As evidenced by the low test errors in the table, the model accurately enhances the accuracy of object detection, contributing to a better overall performance of our methodology.

\begin{table}[ht!]
\centering
\caption{Evaluation results for object location prediction model in the onion-sorting task.}
\setlength{\extrarowheight}{2pt} 
\scalebox{0.85}{ 
\begin{tabular}{c|c|cccc}
\hline
{\multirow{2}{*}{\textbf{Object}}} & {\multirow{2}{*}{\textbf{Axis}}} & \multicolumn{4}{c}{\textbf{Metrics}}\\
& & \textbf{MSE} & \textbf{RMSE} & \textbf{MAE} & \textbf{R\(^2\)} \\
\hline
\multirow{3}{*}{Onion} & X & 0.000203 & 0.012302 & 0.009342 & 0.629628 \\
 & Y & 0.000078 & 0.008898 & 0.006582 & 0.940394 \\
 & Z & 0.000202 & 0.012463 & 0.009058 & 0.712352 \\
\hline
\end{tabular}
}
\label{tab:object_location}
\end{table}

For both tasks, the neuro-symbolic dynamics mapping model is evaluated in three ways. We have calculated the average time it takes to manipulate a single object as well as the average movement jerkiness and the average displacement of the end-effector per object to demonstrate the efficiency of our proposed model in different aspects, particularly focusing on the neuro-symbolic part of our model. The results are shown in Table~\ref{tab:ns_onion} and Table~\ref{tab:ns_pouring}. Figure~\ref{fig:traj} also visually demonstrates that the motions of the proposed method are more closely aligned with human motions than those of the baseline (RRT-connect). More specifically, when the robot uses RRT-Connect, it exhibits erratic stepwise motion between timeslots 3--7, reflecting the algorithm's random sampling and collision check expansions in a restricted environment. This approach seeks any valid joint solution for each incremental step. As a result, it does not inherently preserve smooth or “human-like” trajectories, leading to irregular intermediate steps.  Our neuro-symbolic approach, in contrast, directly encodes human motion patterns and thus generates more fluid transitions.

\begin{table}[ht!]
\fontsize{7.0}{6.5}\selectfont
\addtolength{\tabcolsep}{-4pt}
\setlength\extrarowheight{3pt}
\caption{Comparison of average time (in seconds), joint movement jerkiness (in degrees), and end-effector displacement (in meters) to sort an onion using RRT and our Neuro-Symbolic Dynamics Mapping model.}
\vspace{-2mm}
\begin{center}
\begin{tabular}{c|ccc}
\hline
{\textbf{Method}} & \textbf{Avg. Time} & \textbf{Avg. Jerkiness} & \textbf{Avg. Displacement} \\
 & \textbf{(seconds)} & \textbf{(degrees)} & \textbf{(meters)} \\
\hline
RRT ~\cite{lavalle2001randomized} & 30.87 $\pm$ 2.77 & 914.13 $\pm$ 124.66 & 5.12 $\pm$ 1.28 \\
\textbf{Our Neuro-Symbolic model} & \textbf{14.75 $\pm$ 1.51} & \textbf{264.92 $\pm$ 36.22} & \textbf{2.73 $\pm$ 0.44} \\
\hline
\end{tabular}
\label{tab:ns_onion}
\end{center}
\end{table}

\begin{table}[ht!]
\fontsize{7.0}{6.5}\selectfont
\addtolength{\tabcolsep}{-4pt}
\setlength\extrarowheight{3pt}
\caption{Comparison of average time (in seconds), joint movement jerkiness (in degrees), and end-effector displacement (in meters) to pick a bottle, pour its contents into the designated container, and drop the empty bottle into the bin using RRT-connect and our Neuro-Symbolic Dynamics Mapping model.}
\vspace{-2mm}
\begin{center}
\begin{tabular}{c|ccc}
\hline
{\textbf{Method}} & \textbf{Avg. Time} & \textbf{Avg. Jerkiness} & \textbf{Avg. Displacement} \\
 & \textbf{(seconds)} & \textbf{(degrees)} & \textbf{(meters)} \\
\hline
RRT-connect ~\cite{kuffner2000rrt} & 44.12 $\pm$ 6.21 & 1333.75 $\pm$ 278.15 & 6.48 $\pm$ 2.51 \\
\textbf{Our Neuro-Symbolic model} & \textbf{33.80 $\pm$ 2.89} & \textbf{359.09 $\pm$ 87.78} & \textbf{3.50 $\pm$ 1.02} \\
\hline
\end{tabular}
\label{tab:ns_pouring}
\end{center}
\vspace{-4mm}
\end{table}

As shown in the tables, the results indicate that our model considerably outperforms the baselines in terms of time, smoothness, and power efficiency.

\noindent\textbf{Evaluating \virl{} toward LfO.}
To evaluate the effectiveness of \virl{}, we compare its generated trajectories with another advanced inverse RL algorithm, MAP-BIRL~\cite{choi2011map}. We measure performance using learned behavior accuracy (LBA), which compares the learned policy to the true expert policy. We conducted 10 independent onion sorting trials, each with 30$\pm$5 state-action steps, performed by two different human subjects (5 trials each) to enhance robustness. Predictions from \virl{} achieved an LBA of 90.4\%, compared to 83.3\% that uses discrete state-action pairs as input trajectory. This demonstrates that \virl{} enhances LfO performance by effectively handling continuous state-action values obtained from the human keypoint detection and object localization models. Consequently, \virl{} proves to be highly applicable in real-world robotic domains, leading to substantially better outcomes ({\bf see the supplementary video for more details.})


\section{Challenges and Future Work}
\label{sec:limitations}
During the development of our pipeline, several practical challenges emerged, reflecting the complexities inherent in designing advanced robotic systems. Looking forward, there are promising avenues for further enhancement and expansion.\

\noindent\textbf{Challenges.}
One primary challenge was the discrepancy between the number of joints in cobots and the human arm, which made finding an exact match between joints infeasible. To address this, we fixed certain joint angles in the cobot to prevent further confusion; however, this approach inherently limited the cobot's range of movement capabilities. Another significant challenge was the difference in limb lengths between the cobot and a human, leading to variations in end-effector locations despite similar joint angles. We mitigated this issue by iteratively using the cobot's forward and inverse kinematics models after obtaining a direct mapping of joint angles.\\

\noindent\textbf{Future Work.}
In the future, we plan to extend our model to other manipulation domains and adapt it for cobots with different degrees of freedom. Transferring human dynamics from one domain to another would be another interesting path to explore in the future. These enhancements will broaden the model's applicability and improve its versatility, enabling more accurate and efficient human-like robotic manipulation across various tasks.


\section{Conclusions}
\label{sec:conclusion}

We introduce \virl{}, a novel architecture for generating human-like robotic manipulation from observing humans performing tasks that allow cobots to perform tasks more naturally and smoothly. Designed specifically for reliable robotic systems, our model advances previous methods by leveraging a neuro-symbolic dynamics mapping model, resulting in more efficient robotic maneuvers. Unlike conventional AIRL methods, our approach directly utilizes human body keypoints, facilitating easier adaptation of cobot movements to those of human experts. Our results demonstrate that this approach outperforms the baseline in various aspects, providing an improved solution for teaching cobots to mimic human behavior. Future improvements could include extending the model to other manipulation domains and adapting it for cobots with different degrees of freedom or learning a more generic motion model that could be applied to diverse tasks without the need for retraining.




\begin{acks}
We thank Prasanth Suresh for his valuable experimentation assistance. 
\end{acks}



\bibliographystyle{ACM-Reference-Format} 
\bibliography{adAAMAS25}


\end{document}